\documentclass[10pt,journal,compsoc]{IEEEtran}
\ifCLASSOPTIONcompsoc
  \usepackage[nocompress]{cite}
\else
  \usepackage{cite}
\fi
\ifCLASSINFOpdf
\else
\fi
\usepackage{graphicx}
\usepackage{amsmath,amssymb} 
\usepackage{color}
\usepackage{epsfig}
\usepackage[super]{nth}
\graphicspath{{Figures/Images/}}
\usepackage{tabulary}
\usepackage{subfigure}
\usepackage{capt-of}
\usepackage{ctable}
\usepackage{multirow}

\def\eg{\emph{e.g.}} 
\def\ie{\emph{i.e.}} 
 
\def\etc{\emph{etc.}} 
 
\def\etal{\emph{et al.}}

\begin{document}
%
\title{Skeleton-Based Typing Style Learning For\newline Person Identification}
%
%
%
%

\author{Lior~Gelberg,~David~Mendelovic,~and~Dan~Raviv
\IEEEcompsocitemizethanks{\IEEEcompsocthanksitem L. Gelberg, D. Mendelovic, and D. Raviv are with the Department
of Electrical Engineering, Tel-Aviv University, Tel-Aviv, Israel.\protect\\
E-mail: liorgelberg@mail.tau.ac.il}}

\IEEEtitleabstractindextext{%
\begin{abstract}
We present a novel architecture for person identification based on typing-style, constructed of adaptive non-local spatio-temporal graph convolutional network. Since type style dynamics convey meaningful information that can be useful for\newline person identification, we extract the joints positions and then learn their movements' dynamics. Our non-local approach incre-\newline ases our model's robustness to noisy input data while analyzing joints locations instead of RGB data provides remarkable\newline robustness to alternating environmental conditions,~\eg, lighting, noise,~\etc. We further present two new datasets for typing-\newline style based person identification task and extensive evaluation that displays our model's superior discriminative and generali-\newline zation abilities, when compared with state-of-the-art skeleton-based models.
\end{abstract}

}

\maketitle

\IEEEdisplaynontitleabstractindextext

%
\IEEEpeerreviewmaketitle


%
%
%
%
\begin{figure*}[!h]
\begin{center}
  \includegraphics[width=0.95\linewidth]{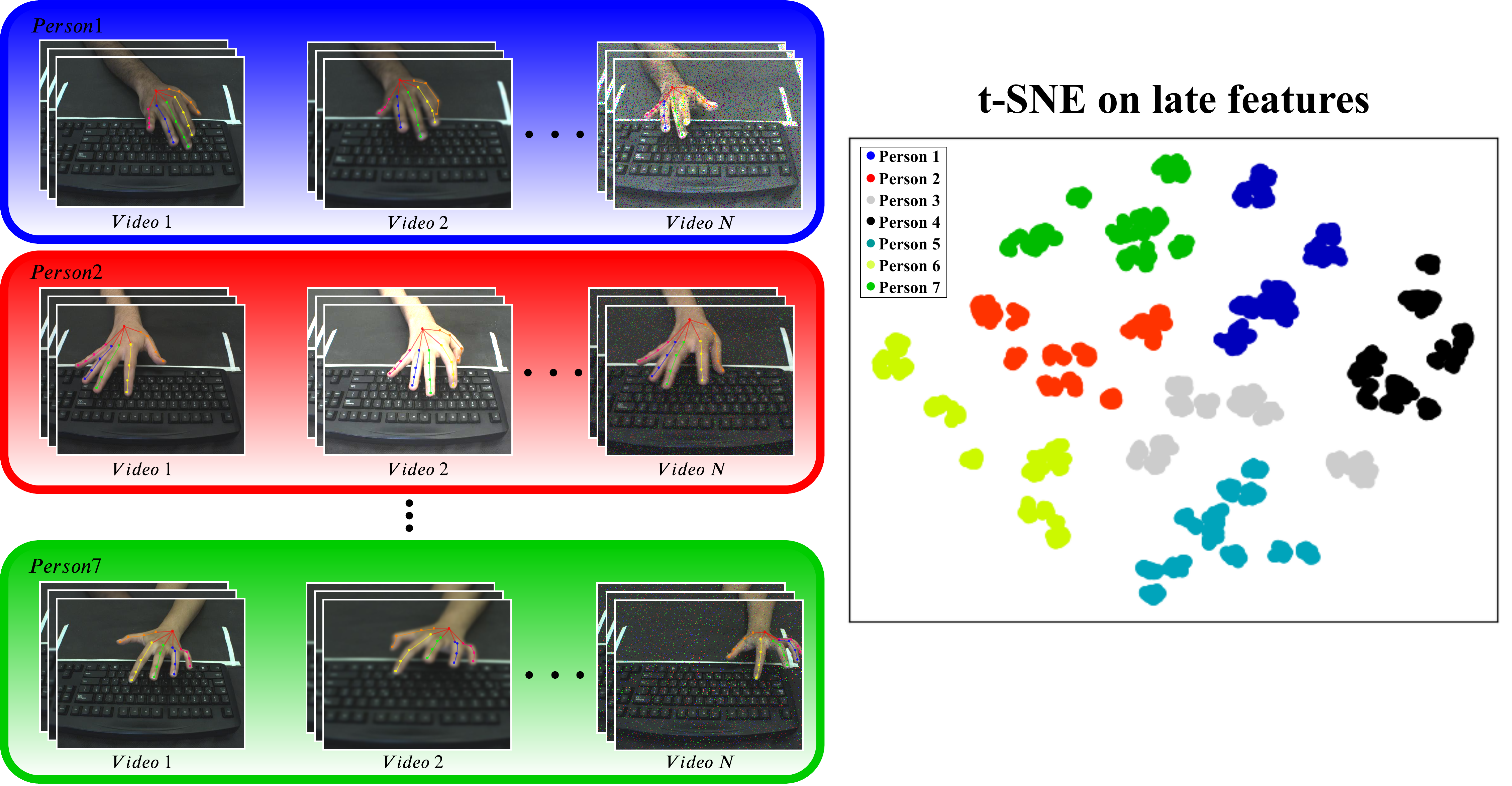}
\end{center}
\caption{t-SNE on late features of 7 out of 60 people appears in~\textit{60Typing10} dataset, where some videos went through data augmentation to simulate changing environmental conditions. Given a video of a person typing a sentence, our model can classify the person according to its unique dynamic,~\ie, typing style, with high accuracy, regardless of scene properties (\eg, lighting, noise,~\etc). The model generalizes the typing style to other sentences, which it never saw during training even when it trains on one sentence type alone, while our non-local approach provides remarkable robustness to noisy data resulting from joints detector failures. Best viewed in color.} 
\label{fig:teaser}
\end{figure*}
\section{Motivation}
\par User identification and continuous user identification are some of the most challenging open problems we face today more than ever in the working-from-home lifestyle due to the COVID-19 pandemic. The ability to learn a style instead of a secret passphrase opens up a hatch towards the next level of person identification, as style is constructed from a person's set of motions and their relations. 
Therefore, analyzing a person's style, rather than rely on its appearance (or some other easily fooled characteristic), can increase the level of security in numerous real-world applications,~\eg, VPN, online education, finance, etc.. 
Furthermore, utilizing a person's style can increase the robustness to changing environmental conditions, as a person's style is indifferent to most scene properties. 
\par Here we focus on a typical daily task - typing as a method for identification and presenting a substantial amount of experiments supporting typing style as a strong indicator of a person's identity. Moreover, our suggested approach makes forgery too complicated, as typing someone's password is insufficient, but typing it in a similar style is needed. Therefore, typing style's remarkable discriminative abilities and high endurance to forgery can offer an elegant and natural solution for both person identification and continuous person identification tasks.

\section{Introduction} \label{INTRODUCTION}
\par Biometrics are the physical and behavioral characteristics that make each one of us unique. Therefore, this kind of character is a natural choice for a person identity verification. Unlike passwords or keys, biometrics cannot be lost or stolen, and in the absence of physical damage, it offers a reliable way to verify someone's identity. Physiological biometrics involves biological input or measurement of other unique characteristics of the body. Such methods are fingerprint \cite{FINGERP1}, blood vessel patterns in the retina \cite{RETINA} and face geometry \cite{FACEGEO,FACENET}. Unlike physiological characteristics, behavioral characteristics encompass both physiological and psychological state. Human behavior is revealed as motion patterns in which their analysis forms the basis for dynamic biometric.
\par Motion analysis is drawing increasing attention due to a substantial improvement in performance it provides in a variety of tasks~\cite{DEEPFEAT1},\cite{VIDPRED}, \cite{JASTPN}, \cite{HANDGESREC}, \cite{VAD}. Motion patterns convey meaningful information relevant to several applications such as surveillance, gesture recognition, action recognition, and many more. These patterns can indicate the type of action within these frames, even manifesting a person's mood, intention, or identity. 
\par Deep learning methods are the main contributors to the performance gain in analyzing and understanding motion that we witness during recent years. 
Specifically, spatio-temporal convolutional neural networks that can learn to detect motion and extract high-level features from these patterns become common approaches in various tasks. Among them, video action classification (VAC), in which given a video of a person performing some action, the model needs to predict the type of action in the video. 
In this work, we take VAC one step further, and instead of trying to predict the action occurs in the input video, we eliminate all action classes and introduce a single action - typing. Now, given a set of videos containing hands typing a sentence, we classify the videos according to the person who is typing the sentence.
\par Over time, researchers in VAC's field presented various approaches, where some use RGB based 2D or 3D convolutions~\cite{DEEPFEAT1,DEEPFEAT5,T3D} while others focus on skeleton-based spatio-temporal analysis~\cite{SB1,RNSSKE1,HCN}. The skeleton-based approach proved its efficiency in cases where the videos are taken under uncontrolled scene properties or in the presence of a background that changes frequently. The skeleton data is captured by either using a depth sensor that provides joint $(x,y,z)$ location or by using a pose estimator such as~\cite{OPENPOSE}, that extracts the skeleton data from the RGB frames. The joint locations are then forwarded to the model that performs the action classification.
\par Recent works in the field of skeleton-based VAC uses architectures of~\textit{Spatio Temporal Graph Convolutional Network} (GCN) as graph-based networks are the most suitable for skeleton analysis since GCN can learn the dependencies between correlated joints. Since Kipf and Welling introduced GCN in their work~\cite{GCN}, other works such as~\cite{GCN3} presented adapted versions of GCN that applied for action classification. These adaptations include spatio-temporal GCN that performs an analysis of both space and time domains as well as adaptive graphs that use a data-driven learnable adjacency matrix. Recently, a two-stream approach~\cite{2sGCN,smart} that is using both joints and bones data is gaining attention. Bones data is a differential version of the joints locations data since it is constructed from subtractions between linked joints. The bones vector contains each bone's length and direction, so analyzing this data is somewhat similar to how a human is analyzing motion. Furthermore, bones can offer new correlated yet complementary data to the joints locations. When combining both joints and bones, the model is provided with much more informative input data, enabling it to learn meaningful information that could not be achieved with a one-stream approach alone.
\par Even though VAC is a highly correlated task to ours, there are some critical differences. The full-body skeleton is a large structure. Its long-ranged joints relations are less distinct than those that appear in a human hand, which has strong dependencies between the different joints due to its biomechanical structure. These dependencies cause each joint's movement to affect other joints as well, even those on other fingers. Thus, when using a GCN containing fixed adjacency matrix, we limit our model to a set of pre-defined connections and not allowing it to learn the relations between joints which are not directly connected. Furthermore, the hand's long-ranged dependencies that convey meaningful information tend to be weaker than the close-range ones, and unless these connections are amplified, we lose essential information.
Our constructed modules are designed to increase vertices and edges inter (non-local) connections, allowing our model to learn non-trivial dependencies and to extract motion patterns from several scales in time, which we refer to as style.
\par In practice, we use a learnable additive adjacency matrix and a non-local operation that increases the long-range dependencies in the layer's unique graph. The spatial non-local operation enables the GCN unit to permute forward better spatial features, and the temporal non-local operation provides the model with a new order of information by generating the inter joints relation in time. Now, each joint interacts with all other joints from different times as well. These dependencies in time help the model gain information regarding the hand and finger posture along time and the typing division among the different fingers. We further apply a downsampler learnable unit that learns to sum each channel information into a single value while causing minimal information loss. As a result, the refined features resulting from the long-ranged dependencies can be reflected as much as possible in the model's final prediction layer. Also, we follow the two-stream approach and apply bones data to a second stream of our model. We train both streams jointly and let the data dictate the relationship between both streams,~\ie, we apply learnable scalars that set each stream's contribution.
\par The final model is evaluated on two newly presented datasets gathered for the task of typing style learning for person identification (person-id). Since this work offers a new task, we present comprehensive comparisons with state-of-the-art skeleton-based VAC models to prove our model's superiority.
The main contributions of our work are in four folds:
\begin{enumerate}
    \item{Develop a Spatio-Temporal Graph Convolution Network (\textit{StyleNet}) for the task of typing style learning which outperforms all compared models in every experiment performed under controlled environmental conditions.}
    \vspace{0.3cm}
    \item{Present substantially better robustness to challenging environmental conditions and noisy input data than all compared state-of-the-art VAC models.}
    \vspace{0.3cm}
    \item{Introduce two new datasets for~\textit{typing style learning for person-id} task.}
    \vspace{0.3cm}
    \item{Introduce an innovative perspective for person-id based on joints locations while typing a sentence.}
\end{enumerate}

\section{Background} \label{RELATED WORK}
AI methods entering the game allow for higher accuracy in various tasks, moving for axiomatic methods towards data-driven approaches. These models focus on the detection of minor changes that were missed earlier by examining dramatically more data. The improvement of hardware allowed us to train deeper networks in a reasonable time and classify in real-time using these complex models. This paper's related works can refer to biometric-based person identification, VAC, Gait recognition, and Gesture recognition. We consider~\textit{style learning} as a biometric-based identification method, and VAC as the motivation for our suggested task. Hence, we discuss these two as related works to ours.
\subsection{Biometrics-based person identification} \label{CLASIC APPROACHES}
Numerous person-identification methods using different techniques and inputs were presented over the years. Ratha \etal \cite{FINGRREC} presented work on fingerprints that uses the delta and core points patterns and ridge density analysis to classify an individual. \cite{KEYST!, KEYST2, KEYST3} studied the use of Keystroke dynamics while others used different biometrics include face recognition \cite{FACEREC, FACEREC2}, iris scan \cite{IRISREC}, and gait analysis \cite{GAITREC}. Identifying a person by his hands was studied by Fong \etal \cite{AUTHHANDGES1}, where they suggested a classification method based on geometric measurements of the user's stationary hand gesture of hand sign language. Roth \etal \cite{TYPEAUTH} presented an online user verification based on hand geometry and angle through time. Unlike \cite{TYPEAUTH}, our method does not treat the hand as one segment but as a deformable part model by analyzing each of the hand joints relations in space and time. Furthermore, our method is more flexible since it is not based on handcrafted features and does not require a gallery video to calculate a distance for its decision.
\subsection{Action recognition}
\par VAC methods are going through a significant paradigm shift in recent years. This shift involves moving from hand-designed features~\cite{HANDFEAT10,HANDFEAT6,HANDFEAT7,HANDFEAT12,HANDFEAT21,HANDFEAT16} to deep neural network approaches that learn features and classify them in an end-to-end manner. Simonyan and Zisserman~\cite{DEEPFEAT1} designed a two-stream CNN  that utilizes RGB video and optical flow to capture motion information. Carreira and Zisserman~\cite{DEEPFEAT5} proposed to inflate 2D convolution layers that pre-trained on ImageNet to 3D, while Diba~\etal~\cite{T3D} presented their inflated DenseNet~\cite{densenet} model with temporal transition layers for capturing different temporal depths. Wang~\etal \cite{DEEPFEAT6} proposed non-local neural networks to capture long-range dependencies in videos. 
\par A different approach for VAC is a skeleton-based method that uses a GCN as well as joints locations as input instead of the RGB video. Yan~\etal~\cite{GCN3} presented their spatio-temporal graph convolutional network that directly models the skeleton data as the graph structure. Shi~\etal~\cite{2sGCN} presented their adaptive graph two-stream model that uses both joints coordinates and bones vectors for action classification and based on the work of~\cite{AGCN} that introduced adaptive graph learning.
\par Inspired by the works presented above, this work follows skeleton-based methods for the task of person-id based on his typing style. Unlike full-body analysis, hand typing style analysis has higher discriminating requirements, which can be fulfilled by better analysis of the hand's global features such as the hand's posture and the fingers intra-relationships as well as inter-relationships in space and time. We claim that all skeleton-based methods presented earlier in this section fail to fulfill these discriminative requirements fully. Therefore, we propose a new architecture that aggregates non-locality with spatio-temporal graph convolution layers. Overall, we explored person-id on seen and unseen sentences under different scenarios.

\section{StyleNet}\label{method}
The human hand is made from joints and bones that dictate its movements. Therefore, to analyze the hand's movements, a Graph Convolutional Network (GCN) is the preferred choice for deep neural network architecture in that case. GCN can implement the essential joints links, sustain the hand's joints hierarchy, and ignore links that do not exist.
\begin{figure}[!h]
\begin{center}
  \includegraphics[width=\linewidth]{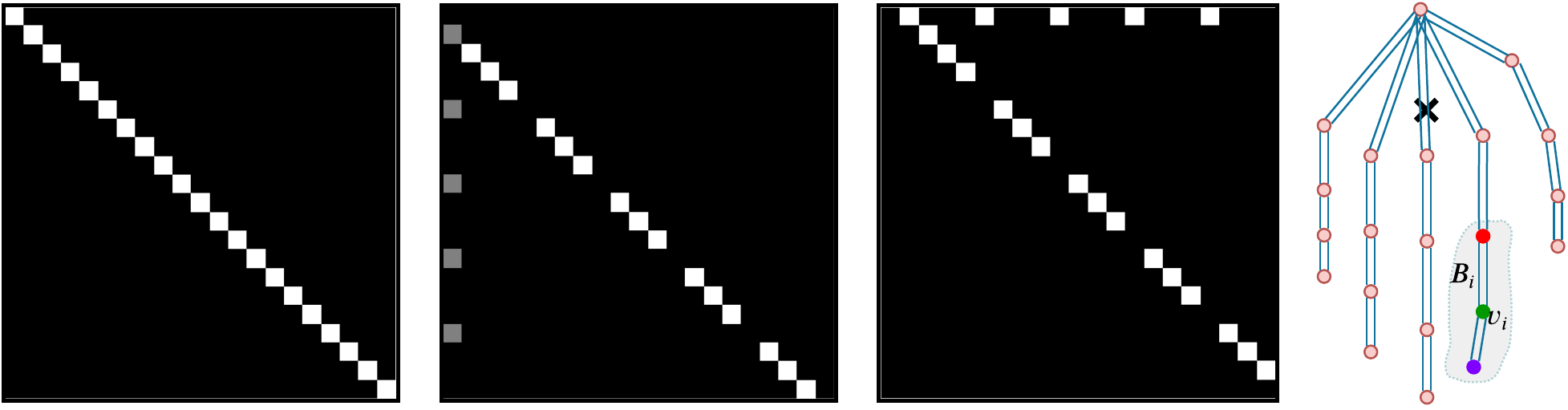}
\end{center}
  \caption{Left to right - adjacency matrix of the $\nth{1}$, $\nth{2}$, and $\nth{3}$ subset, respectively. Right - The hand as a graph. Each circle denotes a joint, and each blue line is a bone connecting two linked joints, i.e., each joint is a vertex, and bones are links in the graph. Black X marks the center of gravity. Gray blob is the subset $B_{i}$ of joint $v_{i}$ and its immediate neighbors. The green joint is $v_{i}$, the joint in red is the immediate neighbor of $v_{i}$ that is closer to the center of gravity, and the joint in purple is the immediate neighbor of $v_{i}$ that is farther from the center of gravity.}
\label{fig:hand_graph}
\end{figure}
\begin{figure*}[!h]
\begin{center}
  \includegraphics[width=0.92\linewidth]{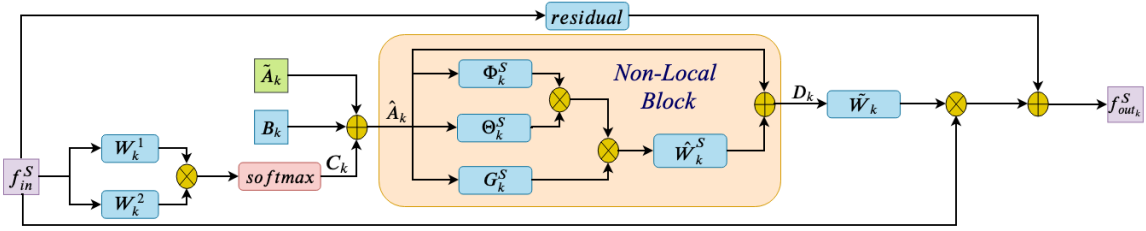}
\end{center}
  \caption{Diagram of our spatial Non-Local GCN unit. Blue rectangles are for trainable parameters. $\bigotimes$ denotes matrix multiplication and $\bigoplus$ denotes element-wise summation. $residual$ block exist only when the unit's $Ch_{in}\neq{Ch_{out}}$. This unit repeated $K_{v}$ times according to the number of subsets, Therefore, $F_{out}^{S} = \sum_{k=1}^{K_{v}}f_{out_{k}}^{S}$.}
\label{fig:spatial_AGCN}
\end{figure*}
\subsection{Spatial Domain}\label{spatial}
Motivated by~\cite{GCN3}, we first formulate the graph convolutional operation on vertex $v_{i}$ as 
\begin{equation} \label{GCN_out}
f_{out}^{S}(v_{i}) = \sum_{v_{j} \in \mathbb{B}_{i}} \frac{1}{Z_{ij}} f_{in}^{S}(v_{j})\cdot w(l_{i}(v_{j})),
\end{equation}
where $f_{in}^{S}$ is the input feature map and superscript $S$ refers to the spatial domain. $v$ is a vertex in the graph and $B_{i}$ is the convolution field of view which include all immediate neighbor $v_{j}$ to the target vertex $v_{i}$. $w$ is a weighting function operates according to a mapping function $l_{i}$. We followed the partition strategy introduced in~\cite{GCN} and construct the mapping function $l_{i}$ as follows: given a hand center of gravity (shown in Figure~\ref{fig:hand_graph}), for each vertex $v_{i}$ we define a set $B_{i}$ that include all immediate neighbors $v_{j}$ to $v_{i}$. $B_{i}$ is divided to 3 subsets, where $B_{i}^{1}$ is the target vertex $v_{i}$, $B_{i}^{2}$ is the subset of vertices in $B_{i}$ that are closer to the center of gravity and $B_{i}^{3}$ is the subset that contains all vertices in $B_{i}$ that are farther from the center of gravity. 
According to this partition strategy, each $v_{j}\in{B_{i}}$, is mapped by $l_{i}$ to its matching subset. $Z_{ij}$ is the cardinality of the subset $B_{i}^{k}$ that contains $v_{j}$. We follow~\cite{Graph1,GCN} method for graph convolution using polynomial parametrization and define a normalized adjacency matrix $A$ of the hand's joints by
\begin{equation} \label{factorization}
\tilde{A} = \Lambda^{-\frac{1}{2}}(A + I)\Lambda^{-\frac{1}{2}}, 
\end{equation}
where $I$ is the identity matrix representing self connections, $A$ is the adjacency matrix representing the connections between joints, and $\Lambda$ is the normalization matrix, where $\Lambda_{ii} = \sum_{j}A_{ij}$. Therefore, $\tilde{A}$ is the normalized adjacency matrix, where its non diagonal elements,~\ie, $\tilde{A}_{ij}$ where $i\neq{j}$ indicate whether the vertex $v_{j}$ is connected to vertex $v_{i}$. Using eq.\ref{GCN_out} and eq.~\ref{factorization} we define our spatial non-local graph convolutional 
(Figure~\ref{fig:spatial_AGCN}) operation as
\begin{equation} \label{ANLGCN_out1}
F_{out}^{S} = \sum_{k}^{K_{v}} W_{k}f_{in}^{S}D_{k},
\end{equation}
where $K_{v}$ is the total number of subsets and is equal to 3 in our case. $W_{k}$ is a set of learned parameters, and $f_{in}^{S}$ is the input feature map. Inspired by \cite{DEEPFEAT6}, we construct $D_{k}$ by  
\begin{equation} \label{ANLGCN_out2}
D_{k} = \hat{W^{S}_{k}}((\Theta^{S}_{k}(\hat{A}_{k}))^{T}\cdot \Phi^{S}_{k}(\hat{A}_{k}))G^{S}_{k}(\hat{A}_{k})) + \hat{A}_{k},
\end{equation}
where superscript $S$ denoted spatial domain. $\Phi^{S}_{k}, \Theta^{S}_{k}$, and $G^{S}_{k}$ are trainable 1D convolutions. These convolutions operate on the graph and embed their input into a lower-dimensional space, where an affinity between every two features is calculated. $\hat{W^{S}_{k}}$ is a trainable 1D convolution used to re-project the features to the higher dimensional space of $\hat{A}_{k}$. We use eq.~\ref{ANLGCN_out2} to apply self-attention on the input signal to enhances the meaningful connections between the features of its input $\hat{A}_{k}$, especially the long-range ones. To construct the input signal $\hat{A}_{k}$, we adopt a similar approach to~\cite{2sGCN} and define $\hat{A}_{k}$ to be
\begin{equation} \label{AGCN_out}
\hat{A}_{k} = \tilde{A}_{k} + B_{k} + C_{k},
\end{equation}
where $\tilde{A}_{k}$ is the normalized adjacency matrix of subset k according to eq.~\ref{factorization}. This matrix is used for extracting only the vertices directly connected in a certain subset of the graph. $B_{k}$ is an adjacency matrix with the same size as $\tilde{A}$ initialized to zeros. Unlike $\tilde{A}_{k}$, $B_{k}$ is learnable and optimized along with all other trainable parameters of the model. $B_{k}$ is dictated by the training data, and therefore, it can increase the model's flexibility and make it more suitable for a specific given task. $C_{k}$ is the sample's unique graph constructed by the normalized embedded Gaussian that calculates the similarity between all vertices pairs according to
\begin{equation} \label{C_eq}
C_{k} = softmax(({W_{k}^{1}f_{in}^{S}})^{T}W_{k}^{2}f_{in}^{S}),
\end{equation}
where $W_{k}^{1}$ and $W_{k}^{2}$ are trainable parameters that embed the input features to a lower-dimensional space, $ softmax$ used for normalizing the similarity operation's output and superscript $S$ denoted spatial domain. $C_{k}$ is somewhat related to $D_{k}$ in the way they both constructed. The main difference is that $C_{k}$ generated by the input features alone, while $D_{k}$ is generated using the input features, the learned adjacency matrix $B_{k}$, and the normalized adjacency matrix $\tilde{A}_{k}$. We use the non-local operation on the addition of $\tilde{A}_{k}$, $B_{k}$ and $C_{k}$ to exploit the information from all three matrices. This information enables the spatial block to permute more meaningful information forward, which contributes to the model's discriminative ability.
\begin{figure*}[!h]
\begin{center}
\includegraphics[width=0.92\linewidth]{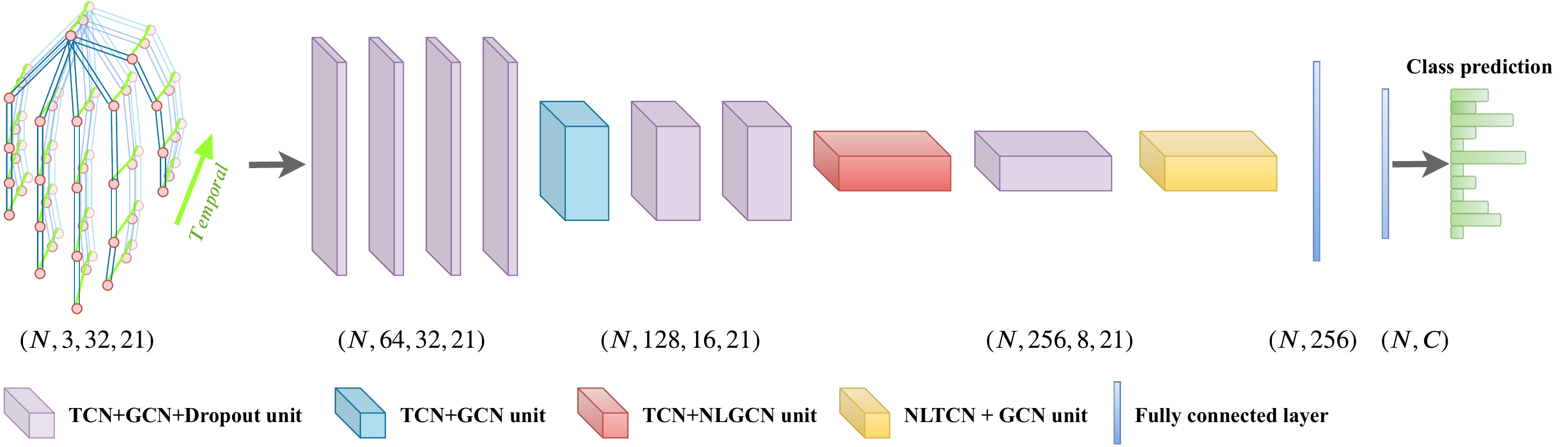}
\end{center}
  \caption{Single stream~\textit{StyleNet} architecture. Input is consists of the 21 coordinates of the hand's joints, while for each joint, we provide a 2D location and a confidence level of its location per frame. The blue lines represent the joints' spatial connections, while the green lines represent the joints' temporal connections. (N,Ch,T,V) Placed under the layers denote Batch size, the number of channels, temporal domain length, and V denotes the joint's index and represents a vertex in the graph, respectively. As for the fully connected layers, N denotes the batch size, and C is the dataset's number of classes.}
\label{fig:architecture}
\end{figure*}
\subsection{Temporal Domain}\label{temporal}
To better exploit the time domain, we place a temporal unit after each spatial GCN unit for better processing longitudinal information. We define $X$ to be $X = Conv(F_{out}^{S})$, where $Conv$ is 2D convolution with kernel size of $9\times1$ and $F_{out}^{S}$ is the spatial unit output. A temporal non-local operation applied on $X$ according to
\begin{equation} \label{ANLGCN_out3}
F_{out}^{\tilde{T}} = W_{{\tilde{T}}}(({\Theta_{\tilde{T}}(X)}^{T}\cdot \Phi_{{\tilde{T}}}(X)) \cdot{G_{{\tilde{T}}}(X)}) + X,
\end{equation}
where ${\tilde{T}}$ denoted the temporal domain. Unlike the spatial non-local operation, here $\Phi_{{\tilde{T}}}, \Theta_{{\tilde{T}}}$, and $G_{{\tilde{T}}}$ are trainable 2D convolutions, since they process the temporal domain and not part of the graph. These convolutions used to embed their input into a lower-dimensional space. Similarly, $W_{{\tilde{T}}}$ is a trainable 2D convolution used to re-project the features to the higher dimensional space of $X$. The temporal non-local operation used for two reasons: First, to better utilize the temporal information regarding the same joint in different places in time. Second, to construct the temporal relations between the different joints through the temporal domain.

\subsection{Downsampling Unit} \label{downsampling}
We further apply a downsampling unit before the classification layer. This unit receives the last temporal unit's output and downsamples each channel into a single value instead of using max or mean pooling. It constructed from~\textit{[fully-connected,batch-normalization,fully-connected]} layers and shared among all channels. The benefit of using this sampling method is that it enables our model to learn summarizing each channel into a single value while minimizing the loss of essential features.

\subsection{Joint decision} \label{JOINT_DECISION}
Encouraged by the work of Shi \etal \cite{2sGCN}, we adopt their two-stream approach and introduce~\textit{StyleNet}. This ensemble model consists of one stream that operates on the joints location, and the other one that operates on the bone vectors. The final prediction constructed according to 
\begin{equation} \label{ensamble}
prediction = \alpha \cdot Output_{Joints} + \beta \cdot Output_{Bones},
\end{equation}
where both $\alpha$ and $\beta$ are trainable parameters that decide on each stream weight for the final prediction. This weighting method increases the model's flexibility since the training data itself determines the weight of each stream. We ensemble the bones data by subtracting pairs of joints coordinates that tied by a connection in the graph. Therefore, the bones data is a differential version of the joints data,~\ie, the high frequencies of the joints data. As deep neural network find it hard to cope with high frequencies, providing a second order of data constructed from these frequencies enable the model to utilize the unique clues hidden in the high frequencies and increase its discriminative ability accordingly.
\section{Implementation details}
\subsection{Data pre-processing} \label{PreProcess}
We used~\textit{YOLOv3}~\cite{YOLOv3} object detector for localizing the hand in the input frame. For the joint detector, we used~\textit{Convolutional Pose Machine}~\cite{CPM} (CPM). This model outputs a belief map of the joints location, where each belief map denotes a specific joint. The joint's location is given by a Gaussian whose $\sigma$ and peak value are set according to the model's confidence,~\ie, small $\sigma$ with large peak value if the model is very confident in the location of the joint and large $\sigma$ with small peak value otherwise. In that manner, the CPM model can predict a location for a joint, even when the joint is entirely or partially occluded in a given frame. It can predict the joint's location according to the hand's context and decrease its belief score in exchange. This kind of method can help with cases of hidden joints since~\textit{StyleNet} can utilize the joint's score as an indicator for the liability of the data related to that joint.

\subsection{Models implementation details} \label{IMPLENTATION}
\par {\bf Pre-process pipeline:} We implemented our models for the pre-process using Tensorflow framework. An Input frame of size $240\times320$ was given to the hand localizer to output a bounding box coordinates of the hand in the given frame. We cropped the hand centered frame according to the given bounding box and resized the cropped frame to a size of $368\times368$ with respect to the aspect ratio. The resized frame is given to the joint detector that produces belief maps in return. The belief maps are resized back to fit the original frame size with respect to the translation of the bounding box produced by the hand localizer. Finally, $argmax$ is applied to each belief map to locate the joints coordinates. We repeat this process for the entire dataset to produce the joints locations matrix, which consists of all 21 joints locations and belief scores by frame.
\par {\bf StyleNet:} We implemented StyleNet using PyTorch framework. We defined $A$ which is the adjacency matrix of the hand's joints and normalized it according to eq.~\ref{factorization}, where $\Lambda^{ii}_{k} = \sum_{j}(A_{k}^{ij}) + \sigma$ and $\sigma$ equal to 0.001 is used to avoid empty rows. For each video, we sample a total of 32 matrices, where each matrix refers to a certain frame and comprises the frame's 21 $(x,y)$ joints locations and their belief score. We created the bone data by subtracting the $(x,y)$ coordinates of each neighboring joints pair to extract the bone vectors, while we multiplied both neighboring joints belief score to produce a bone belief score. Our model (figure~\ref{fig:architecture}) is following the AGCN~\cite{2sGCN} architecture, where each layer constructed from a spatial GCN unit that processes the joints or bones intra-frame relations and a temporal unit that process the temporal inter-frame relations. The model's~\nth{8} GCN unit modified according to eq.~\ref{ANLGCN_out1} to improve the long-range dependencies of the spatial feature maps before expanding the number of feature maps channels. We also modify the~\nth{10} TCN unit according to eq.~\ref{ANLGCN_out3} to improve the long-range dependencies between the different frames. The downsampling unit is applied after the~\nth{10} TCN unit for better downsampling of the final feature maps before forwarding to the classification layer.

\subsection{Training details} \label{TRAINING}
\par {\bf Pre-process:} We used~\textit{YOLOv3} model pre-trained on~\textit{COCO} dataset~\cite{COCO}. To train the model for our task, we created a single "hand" label and used~\textit{Hands} dataset~\cite{HANDS} that contains $\sim13k$ training and $\sim2.5k$ validation images, labeled with hands bounding boxes location. We used Adam optimizer with an initial learning rate of 1e-3 and ran our training with a batch size of 16 for 150 epochs. We trained~\textit{CPM} model using trained weights~\cite{CPM_GIT} as an initial starting point. We used 1256 random frames from our~\textit{80Typing2} dataset labeled with their joints locations. Training data consist of 1100 frames and 156 frames used for validation. Data augmentation applied during training to prevent overfitting. We used Adam optimizer with an initial learning rate of 1e-3 and a batch size of 16 for a total of 960 epochs.
\par {\bf StyleNet:} We used a batch size of 32, where each sampled video consists of 32 sampled frames from the entire video. We used Adam optimizer with an initial learning rate of 1e-3, a momentum of 0.9, and a weight decay of 1e-5. Both stream weights initialized to 1. A dropout rate of 0.3 was applied to increase the model's generalization ability. We trained the model for 100 epochs and decreased the learning rate by a factor of 10 after 40, 70, and 90 epochs. No data augmentation needed due to the natural augmentation of the data results from the sampling of the video.
\begin{table*}[t]
    \caption{Test accuracy of user classification on unseen sentences on~\textit{60Typing10}. $[\alpha,\beta,\gamma]$ denotes the number of sentences for train, validation and test, respectively}
    \label{60Ty10}
    \footnotesize
        \begin{tabular}[t]{p{2.7cm} c}
            \multicolumn{2}{c}{\textbf{[4,2,4]}}\\
            \hline
            Model & Acc(\%)\\
            \hline
            HCN~\cite{HCN} & 91.98\\
            STGCN~\cite{GCN3} & 97.09\\
            3sARGCN~\cite{ARGCN} & 95.8\\
            PBGCN~\cite{PBGCN} & 98.9\\
            2sAGCN~\cite{2sGCN} & 99.04\\ 
            \hline
            StyleNet & \textbf{99.84}\\
            \hline   
        \end{tabular}
        \hfill
        \begin{tabular}[t]{p{2.7cm} c}
            \multicolumn{2}{c}{\textbf{[3,2,5]}}\\
            \hline
            Model & Acc(\%)\\
            \hline
            HCN~\cite{HCN} & 84.16\\
            STGCN~\cite{GCN3} & 97.21\\
            3sARGCN~\cite{ARGCN} & 93.6\\
            PBGCN~\cite{PBGCN} & 98.6\\
            2sAGCN~\cite{2sGCN} & 98.82\\ 
            \hline
            StyleNet & \textbf{99.77}\\
            \hline   
        \end{tabular}
        \hfill
        \begin{tabular}[t]{p{2.7cm} c}
            \multicolumn{2}{c}{\textbf{[2,2,6]}}\\
            \hline
            Model & Acc(\%)\\
            \hline
            HCN~\cite{HCN} & 79.53\\
            STGCN~\cite{GCN3} & 94.94\\
            3sARGCN~\cite{ARGCN} & 91.35\\
            PBGCN~\cite{PBGCN} & 96.94\\
            2sAGCN~\cite{2sGCN} & 97.97\\ 
            \hline
            StyleNet & \textbf{99.5}\\
            \hline   
        \end{tabular}
\end{table*}

\section{Experiments} \label{EXPERIMENT}
\par Since there is no dataset for the suggested task, we created~\textit{80Typing2} and~\textit{60Typing10} datasets for the evaluation of our model. We compared our model with skeleton-based action classification models using the new datasets under various test cases, simulating user identification, and continuous user identification tasks. In~\ref{dataset} we present our new datasets and our main experiments results presented in~\ref{classification} and~\ref{re-id}. We further compare our model under challenging scenarios such as noisy input data~\ref{noisy} and presents our chosen skeleton-based approach superiority over RGB modality in~\ref{VsRGB}. In \ref{2DVs3D}, we provide an additional comparison between the models using 3D input data taken from~\textit{How We Type} dataset~\cite{HWT}.
\par In all experiments, we split our data between train, validation, and test sets randomly according to the experiment's settings for an accurate evaluation of the models. Each input video consists of 32 sampled frames from the entire video. We tested each trained model for tens of times and set its accuracy according to all tests' mean accuracy. It is crucial to evaluate each trained model several times since we sample only 32 frames and not use the entire video.

\subsection{80Typing2 and 60Typing10 datasets} \label{dataset}
\par We present two new datasets created for~\textit{typing style learning for person identification} task. The datasets recorded using a simple RGB camera with 100 fps for~\textit{80Typing2} and 80 fps for~\textit{60Typing10}. No special lighting used, and the camera's position remained fixed through all videos. No jewelry or any other unique clues appear in the videos. Both men and women, as well as right and left-handed, appear in the dataset. All participants were asked to type the sentences with their dominant hand only. 
\par~\textit{80Typing2} dataset consists of 1600 videos of 80 participants. Each participant typed two different sentences, while each sentence repeated ten times. This setting's main purpose is simulating a scenario where a small number of different sentences, as well as many repetitions from each sentence, are provided. As each person encounters a changing level of concentration, typing mistakes, distractions, and accumulate fatigue, the variety in the typing style of each participant revealed among a large number of repetitions of each sentence. Therefore, this dataset deals with a classification of a person under intra-sentence varying typing style,~\ie, changing motion patterns of the same sentence, and inter-person changing level of typing consistency. Additionally, this dataset can suggest a scenario where a model learns on one sentence and need to infer to another sentence it never saw during training. 
\par~\textit{60Typing10} dataset consists of 1800 videos of 60 participants. Each participant typed ten different sentences, while each sentence repeated three times. Unlike~\textit{80Typing2},~\textit{60Typing10} setting's purpose is simulating a scenario where a large number of different sentences, as well as a small number of repetitions from each sentence, are provided. The large abundance of different sentences,~\ie, different motion patterns, reveals each participant's unique typing style, while the small amount of repetitions supports each participant variance in the typing style. Therefore, this dataset deals with classification of a person under inter-sentence varying motion patterns, and in order for the model to generalize well to sentences it never saw during training, it must learn to classify each person by his unique typing style,~\ie, learn to classify the different people according to their unique typing style. 
\par We labeled 1167 random frames from~\textit{80Typing2} with their corresponding joints location to train a joint detector.

\subsection{User classification on unseen sentences}\label{classification}
In this experiment, we simulate a test case of continuous user identification by testing our model's ability to infer on unseen sentences,~\ie, different motion patterns. We split our data by sentence type and let the model train on a certain set of sentences while testing performed on a different set of sentences which the model never saw during training,~\ie, different types of sentences the user typed. Therefore, to perform well, the model must learn the unique motion style of each person. 
\par The experiment performed on~\textit{60Typing10} in the following manner, we split our data in three ways, wherein each split a different number of sentences is given for training. We randomly split our data by sentences to train, validation, and test sets according to the split settings. We applied the same division to all other models for legitimate comparison. For~\textit{80Typing2}, we randomized the train sentence, and the other sentence divided between validation and test where two repetitions were used for validation and eight for test.
\par Results for this experiment on~\textit{60Typing10} and~\textit{80Typing2} appears in table~\ref{60Ty10} and~\ref{80Tab1}, respectively. Our model outperforms all other compared models by an increasing margin as less training sentences are provided, which indicates our model's superior generalization ability. 

\subsection{User classification on seen sentences}\label{re-id}
In this experiment, we simulate a test case of user identification (access control by password sentence) by testing our model's ability to infer the same movement patterns,~\ie, sentences, he saw during training and other repetitions of these patterns. We use a large number of sentence repetitions to test the robustness to the variance in the typing style by simulating a scenario where a small amount of different motion patterns,~\ie, sentence type, is given along with a substantial variance in these patterns resulting from a large number of repetitions. 
\par This experiment is performed by dividing~\textit{80Typing2}'s ten repetitions of each sentence as follows: five for train, one for validation, and four for test. We trained each model on the train set and tested its accuracy on the seen sentences but unseen repetitions. 
\par According to the experiment's results, which appears in table~\ref{Tabreid}, it is clear that this specific task is not complex and can be addressed by other methods. However, it proves that our models' extra complexity does not harm the performance in the simpler "password sentence" use cases.
\begin{table}[h]
        \begin{minipage}[t]{0.45\linewidth}
        \centering
        \captionof{table}{Test accuracy of user classification on~\textbf{unseen} sentence experiment on~\textit{80Typing2}. Each model training set is constructed from one sentence, while validation and test sets constructed from the other sentence that did not appear in the training}
        \begin{tabular}[t]{l c}
            \hline
            Model & Acc(\%)\\
            \hline
            HCN~\cite{HCN} & 94.18\\
            STGCN~\cite{GCN3} & 93.59\\
            3sARGCN~\cite{ARGCN} & 91.08\\
            PBGCN~\cite{PBGCN} & 95.98\\
            2sAGCN~\cite{2sGCN} & 96.88\\ 
            \hline
            StyleNet & \textbf{99.57}\\
            \hline
                    \end{tabular}
                    \label{80Tab1}
        \end{minipage} \qquad
        \begin{minipage}[t]{0.45\linewidth}
        \centering
        \captionof{table}{Test accuracy of user classification on~\textbf{seen} sentences experiment on~\textit{80Typing2}. The training set includes five repetitions from both sentences, while the validation and test sets include one and four repetitions from both sentences, respectively}
        \label{Tabreid}
        \begin{tabular}[t]{l c}
            \hline
            Model & Acc(\%)\\
            \hline
            HCN~\cite{HCN} & 99.66\\
            STGCN~\cite{GCN3} & 99.64\\
            3sARGCN~\cite{ARGCN} & 99.44\\
            PBGCN~\cite{PBGCN} & 99.84\\
            2sAGCN~\cite{2sGCN} & 99.85\\ 
            \hline
            StyleNet & \textbf{99.98}\\
            \hline
        \end{tabular}
        \end{minipage}
    \end{table}

\subsection{Noisy data} \label{noisy}
The skeleton-based approach is dependent on a reliable joints detector that extracts the joint's location from each input frame. To challenge our model, we experimented with a scenario similar to~\ref{classification} (the more challenging task simulating continuous user identification), where during inference, the joints detector is randomly failing and providing noisy data,~\ie, incorrect joints location. 
\par We performed this experiment by training all models as usual, while during test time, we randomly zeroed $(x,y, score)$ data of a joint. The amount of joints that zeroed is drawn uniformly among [0,1,2], while the decision of which joint values to zero is random, but weighted by each joint tendency to be occluded,~\eg, the tip of the thumb's joint has a higher probability of being drawn than any of the ring fingers which tend less to be occluded while typing. 
\begin{table}[b]
\centering
\caption{Test accuracy for noisy data experiment on~\textit{60Typing10}. Training conducted as usual, but during test time, we randomly zeroed joint $(x,y,score)$ to simulate a situation where the data is noisy or some joint's location is missing. $[\alpha,\beta,\gamma]$ denotes the number of sentences given for train, validation, and test, respectively}
\label{Tabab1}
\begin{tabular}{l c c c}
\hline
Model & [4,2,4] Acc(\%) & [3,2,5] Acc(\%) & [2,2,6] Acc(\%)\\
\hline
HCN~\cite{HCN} & 57.87 & 53.46 & 45.06\\
STGCN~\cite{GCN3} & 70.03 & 68.3 & 60.61\\
3sARGCN~\cite{ARGCN} & 71.36 & 69.35 & 67.92\\
PBGCN~\cite{PBGCN} & 83.96 & 82.75 & 80.4\\
2sAGCN~\cite{2sGCN} & 73.33 & 71.34 & 68.83\\ 
\hline
StyleNet & \textbf{91.79} & \textbf{87.57} & \textbf{85.24}\\
\hline
\end{tabular}
\end{table}
\begin{table*}[!h]
\centering
\caption{Test accuracy for uncontrolled environment experiment on~\textit{60Typing10}. RGB models trained with data augmentation while during test time, a different set of augmentations applied. $[\alpha,\beta,\gamma]$ denotes the number of sentences for train, validation, and test, respectively. env. denotes environment}
    \label{Tabab2}
\begin{tabular}{l c c c c c c}
\hline
\multirow{2}{*}{Model} & \multicolumn{2}{c}{\textbf{\underline{[4,2,4] Acc(\%)}}} & \multicolumn{2}{c}{\textbf{\underline{[3,2,5] Acc(\%)}}} & \multicolumn{2}{c}{\textbf{\underline{[2,2,6] Acc(\%)}}}\\
        & Controlled env. & Uncontrolled env. & Controlled env. & Uncontrolled env. & Controlled env. & Uncontrolled env.\\
        \hline
I3D~\cite{DEEPFEAT5} & 99.68 & 63.12 & 99.75 & 59.16 & \textbf{99.7} & 62.30\\
T3D~\cite{T3D} & 98.85 & 56.89 & 99.01 & 54.67 & 98.64 & 54.06\\
\hline
StyleNet & \textbf{99.84} & \textbf{99.59} & \textbf{99.77} & \textbf{99.57} & 99.5 & \textbf{99.17}\\
\hline
\end{tabular}
\end{table*}
\begin{table*}[!h]
\begin{center}
\caption{Test accuracy of user classification on unseen sentences on~\textit{How We Type} using 3D input data. $[\alpha,\beta,\gamma]$ denotes the number of sentences for train, validation and test, respectively}
\label{HWT_res}
\begin{tabular}{l c c c c c}
\hline
\multirow{1}{*}{Model} & [5,10,35] Acc(\%) & [10,10,30] Acc(\%) & [15,10,25] Acc(\%) & [20,10,20] Acc(\%) & [25,10,15] Acc(\%)\\
\hline
HCN~\cite{HCN} & 92.46 & 96.27 & 97.3 & 98.32 & 98.82\\
STGCN~\cite{GCN3} & 95.72 & 97.92 & 98.24 & 98.79 & 98.96\\
3sARGCN~\cite{ARGCN} & 94.7 & 97.76 & 98.08 & 98.56 & 98.89\\
PBGCN~\cite{PBGCN} & 98.51 & 99.07 & 99.48 & \textbf{99.61} & 99.7\\
2sAGCN~\cite{2sGCN} & 97.75 & 98.33 & 98.73 & 98.96 & 99.01\\ 
\hline
StyleNet & \textbf{99.46} & \textbf{99.48} & \textbf{99.51} & 99.58 & \textbf{99.79}\\
\hline
\end{tabular}
\end{center}
\end{table*}
\begin{table*}[!h]
\begin{center}
\caption{Test accuracy of user classification on unseen sentences on~\textit{60Typing10} when using 3D or 2D input data. $[\alpha,\beta,\gamma]$ denotes the number of sentences for train, validation, and test, respectively.}
\label{HWT_2D_3D}
\begin{tabular}{l c c c c c}
\hline
\multirow{1}{*}{Model} & [5,10,35] Acc(\%) & [10,10,30] Acc(\%) & [15,10,25] Acc(\%) & [20,10,20] Acc(\%) & [25,10,15] Acc(\%)\\
\hline
StyleNet 2D & 99.41 & 99.47 & \textbf{99.54} & \textbf{99.59} & 99.78\\
StyleNet 3D & \textbf{99.46} & \textbf{99.48} & 99.51 & 99.58 & \textbf{99.79}\\
\hline
\end{tabular}
\end{center}
\end{table*}
\par According to the experiment's results in table~\ref{Tabab1}, our model is much more robust to noisy data. The non-local approach helps the model rely less on a particular joint and provides a more global analysis of each person's typing style, which increases the model's robustness in cases of noisy data.

\subsection{Uncontrolled environment} \label{VsRGB}
In this experiment, we compared our method with VAC RGB-based methods in an uncontrolled environment scenario. 
Even though RGB based methods perform well in a controlled environment, their performance tends to decrease severely under alternating scene properties such as lighting and noise. Even though data augmentation can increase these methods robustness to challenging environmental conditions, it is impossible to simulate all possible scenarios. Therefore, using an RGB-based approach in real-world scenarios tends to fail in the wild. Therefore, we explored our method's robustness under challenging environmental conditions to verify the skeleton-based approach superiority in the task of~\textit{typing style learning for person identification}. 
\par We performed this experiment in a similar manner to~\ref{classification}, but with some differences. We trained each model using data augmentation techniques such as scaling, lighting, and noise. Later, during test time, we applied different data augmentations,~\eg, different lighting, and noise models, than those used during training on the input videos.

\par Results for this experiment appear in table~\ref{Tabab2}. While all the compared methods achieved a high accuracy rate under a controlled environment, their accuracy rate dropped in an uncontrolled environment scenario. Our method's performance did not change except for a slight decline of less than 0.5\% in its accuracy rate. It is much easier to train a joint detector to operate in an uncontrolled environment since it locates the joints by the input image and the hand context altogether. Unlike the image appearance, the hand context is not dependent on the environment. Therefore, the joints localizer can better maintain its performance under varying conditions, making our pipeline resilient to this scenario.
\subsection{2D Vs. 3D data}\label{2DVs3D}
We conducted an experiment that evaluates our model using a 3D input and the trade-off between 3D and 2D input data.
\par We used How We Type dataset~\cite{HWT} that contains 3D coordinates of 52 joints from both hands and a total of 30 different persons, where each person typed 50 sentences. Overall, we tested five different splits of the data, where each split contains a different number of training sentences. We randomly divided the data between training, validation, and test in a similar manner to~\ref{classification} according to the partitioning setting of each split. We repeated this scheme several times for an accurate assessment of the model's performance.
\begin{figure}[!h]
\centering  
\subfigure[$\nth{1}$ subset]{\includegraphics[width=0.32\linewidth]{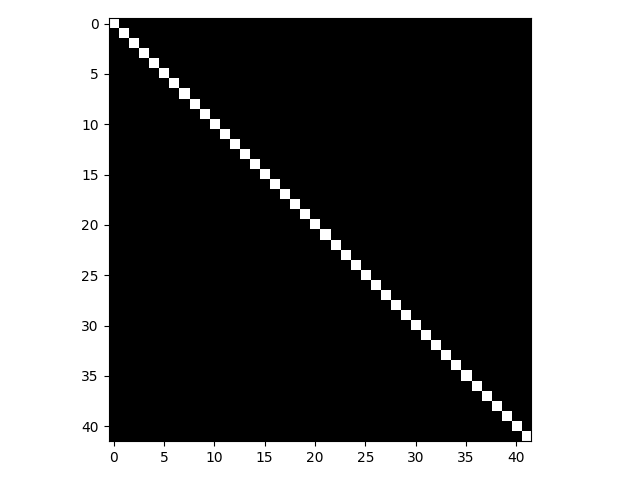}}
\subfigure[$\nth{2}$ subset]{\includegraphics[width=0.32\linewidth]{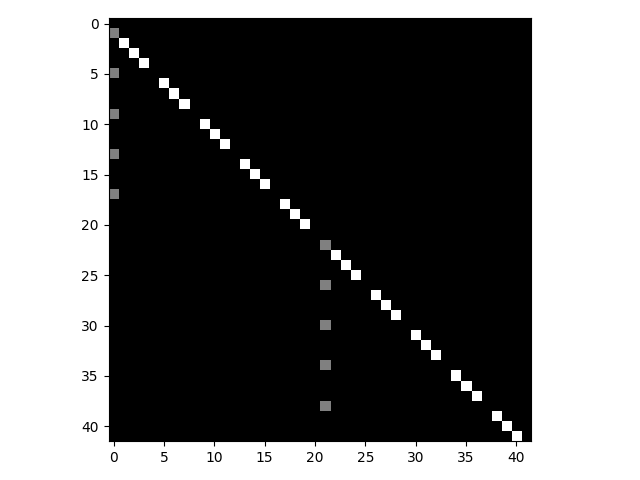}}
\subfigure[$\nth{3}$ subset]{\includegraphics[width=0.32\linewidth]{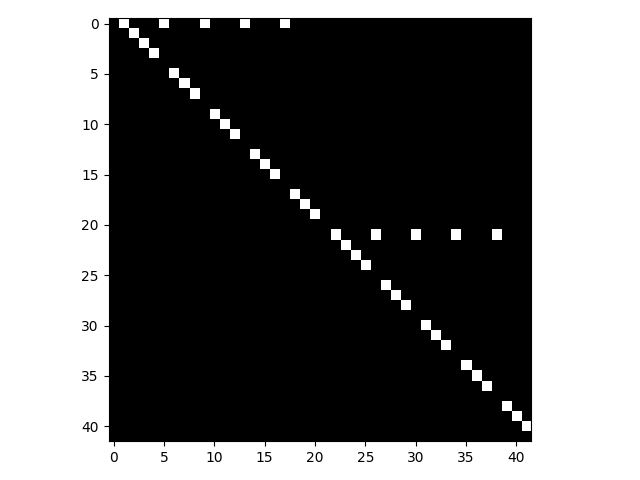}}
\caption{Adjacency matrices of two hands. Each matrix is built by diagonally concatenating two replicas of its one-hand version.}\label{Two_hands_graph}
\end{figure}
\par We used 21 out of 26 joints for each hand for consistency with all other experiments and followed~\cite{GCN3} partition strategy, which was mentioned in the paper. Figure~\ref{Two_hands_graph} contains the adjusted adjacency matrix that enables our model to learn the unique dependencies between the joint of both hands. When we tested our model with 3D coordinates as input, $z$ axis data replaced the $score$ input. Therefore, each frame data consist of 42 $(x,y,z)$ coordinates of joints from both hands.
\par The results for this experiment appear in table~\ref{HWT_res}, where we can see that even though our model trained on only 10\% of the entire data, it achieved a high accuracy rate and outperformed all other models. Results for the trade-off between 2D and 3D input data appear in table~\ref{HWT_2D_3D}. According to the results, we can see that our model achieves similar performance when provided either with 2D or 3D input data. Unlike other tasks where the model benefits from the~\nth{3} dimension, it seems unneeded in this task.

\section{Ablation Study}
We conducted an ablation study to examine the effectiveness of our added blocks using~\textit{60Typing10}. We performed this experiment in the same manner as~\ref{classification}, as this scenario offers a more challenging test case in which the true value of our comprised modules can manifest. The models training was conducted as described in section~\ref{TRAINING}. 
\par According to the results reported in table~\ref{module_comp}, we can see that each added block improves the accuracy rate when compared with the baseline. The most significant improvement was achieved when all the blocks added together. 
On a broader note, applying \cite{GCN3}, \cite{2sGCN}, or any other variant of these methods on a small deformable structure will bias toward close-ranged dependencies (due to the~\textit{Softmax} normalization constructing $C_{k}$). As the close and long-range concept is no longer applicable in our task (moving only one of the hand's joints is almost impossible), these models achieve inferior results to our model, which focuses on non-local spatial and temporal connectivity. Specifically, it constructs a new order of information. Each joint can interact with~\textbf{all} relevant (by attention) joints from all time steps, helping our model extract more meaningful motion patterns in space and time.

\begin{table}[ht]
\begin{center}
\caption{Test accuracy of user classification on unseen sentences on~\textit{60Typing10} when adding each module to our baseline. $[\alpha,\beta,\gamma]$ denotes the number of sentences for train, validation, and test, respectively. NL denotes non-local, SNL denotes temporal non-local unit, and SNL denotes spatial non-local unit}
\label{module_comp}
\begin{tabular}{l c c c}
\hline
\multirow{2}{*}{Model} & [4,2,4] & [3,2,5] & [2,2,6]\\
                       & Acc(\%) & Acc(\%) & Acc(\%) \\
\hline
AGCN~\cite{2sGCN} & 99.04 & 98.82 & 97.97\\
\specialrule{.3em}{.2em}{.2em}
W downsample unit & 99.47 & 99.39 & 98.72\\
W downsample + TNL & 99.62 & 99.59 & 99.13\\
W downsample + SNL & 99.76 & 99.71 & 99.28\\
\specialrule{.2em}{.1em}{.1em}
StyleNet & \textbf{99.84} & \textbf{99.77} & \textbf{99.50}\\
\hline
\end{tabular}
\end{center}
\end{table}

\section{Conclusions} \label{conclusion}
We introduced~\textit{StyleNet}, a novel new architecture for skeleton-based typing style person identification. Motivated by~\cite{DEEPFEAT6}, we redesigned the spatial-temporal relationships allowing for a better longitudinal understanding of actions.~\textit{StyleNet} evaluated on the newly presented~\textit{80Typing2} and~\textit{60Typing10} datasets and outperformed all compared skeleton-based action classification models by a large margin when tested in the presence of noisy data and outperformed when tested under controlled conditions.

\ifCLASSOPTIONcaptionsoff
  \newpage
\fi



%

\bibliographystyle{IEEEtran}
\bibliography{egbib}


%

\begin{IEEEbiography}[{\includegraphics[width=1in,height=1.25in,clip,keepaspectratio]{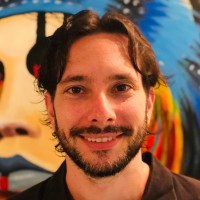}}]{Lior Gelberg}
is a Ph.D. candidate in the Electrical Engineering department at Tel-Aviv University. He received his BSc in 2016 and his MSc in 2020, both in Electrical Engineering from Tel-Aviv University. His main areas of interest are Motion and Gesture recognition, Video analysis, and forensics.
\end{IEEEbiography}

\begin{IEEEbiography}[{\includegraphics[width=1in,height=1.25in,clip,keepaspectratio]{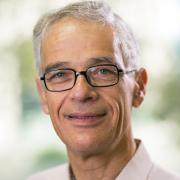}}]{David Mendelovic}
is a professor of Electrical Engineering at Tel-Aviv University. He held a post-doctoral position at the University of Erlangen-Nurnberg, Bavaria as a MINERVA Postdoctoral Fellow.  He is the author of more than 130 refereed journal articles and numerous conference presentations. His academic interests include optical information processing, signal and image processing, diffractive optics, holography, temporal optics, optoelectronic and optically interconnected systems.
\end{IEEEbiography}


\begin{IEEEbiography}[{\includegraphics[width=1in,height=1.25in,clip,keepaspectratio]{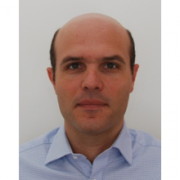}}]{Dan Raviv}
is a senior lecturer of Electrical Engineering at Tel-Aviv University. He held a post-doctoral position at Massachusetts Institute of Technology. He has worked in various areas of image and shape analysis in computer vision, image processing, and computer graphics. Raviv’s main areas of interest are machine learning problems with geometric flavor, medical imaging and robotics.
\end{IEEEbiography}




\end{document}